\documentclass[a4paper,twoside]{article}

\usepackage{epsfig}
\usepackage{subcaption}
\usepackage{calc}
\usepackage{amstext}
\usepackage{amsthm}
\usepackage{multicol}
\usepackage{pslatex}
\usepackage{apalike}
\usepackage{SCITEPRESS}     

\usepackage{epstopdf}
\usepackage{fourier} 
\usepackage{array}
\usepackage{makecell}
\usepackage[export]{adjustbox}

\usepackage{tabularx}
\usepackage{diagbox}
\usepackage{slashbox}
\usepackage{amsmath,amssymb,amsfonts}
\usepackage{graphicx}
\usepackage{textcomp}
\usepackage{xcolor}
\def\BibTeX{{\rm B\kern-.05em{\sc i\kern-.025em b}\kern-.08em
		T\kern-.1667em\lower.7ex\hbox{E}\kern-.125emX}}

\usepackage{times}
\usepackage{latexsym}
\usepackage{caption}
\usepackage{algorithm}
\usepackage{algorithmic}

\usepackage{multirow}
\usepackage{booktabs}
\usepackage{tabularx}
\usepackage{enumitem}  

\usepackage{balance}

\begin{document}

\title{Predicting the intended action using internal simulation of perception}

\author{\authorname{Zahra Gharaee}
\affiliation{Computer Vision Laboratory (CVL), Department of Electrical Engineering, \\ University of Link\"oping, 581 83 Link\"oping, Sweden}
\email{zahra.gharaee@liu.se}
}


\abstract{This article proposes an architecture, which allows the prediction of intention by internally simulating perceptual states represented by action pattern vectors. To this end, associative self-organising neural networks (A-SOM) is utilised to build a hierarchical cognitive archi- tecture for recognition and simulation of the skeleton based human actions. The abilities of the proposed architecture in recognising and predicting actions is evaluated in experiments using three different datasets of 3D actions. Based on the experiments of this article, apply- ing internally simulated perceptual states represented by action pattern vectors improves the performance of the recognition task in all experiments. Furthermore, internal simulation of perception addresses the problem of having limited access to the sensory input, and also the future prediction of the consecutive perceptual sequences. The performance of the system is compared and discussed with similar architecture using self-organizing neural networks (SOM).}

\onecolumn \maketitle \normalsize \setcounter{footnote}{0} \vfill

\section{Introduction}
\vspace{-0.5em}
For efficient Human-Robot-Interaction, it is important that the robot can predict the behaviour of the human, at least for the nearest future. In Human-Human interaction we do this by reading the intentions of others. This is done using our capacity for mind reading, intention recognition is one of the core processes of mind reading \cite{bonchek2014towards}, where we simulate the thoughts of others. We can predict about behavior and mental states of another person based on our own behavior and mental states if we were in his/her situation. This occurs by simulating another person’s actions and the stimuli he/she is experiencing using our own behavioral and stimulus processing mechanisms \cite{breazeal2005learning,bonchek2014towards}. The overarching question of this article is to develop a computational model of an artificial agent for intention prediction, the ability to extend an incomplete sequence of actions to its most likely intended goal.

\subsection{Theoretical background}
\vspace{-0.5em}
The perceptual processes normally elicited by some ancillary input can be mimicked by the brain \cite{hesslow2002conscious}. This relates to an idea that higher organisms are capable of simulating perception, which is supported by a large number of evidences. As shown by several neuroimaging experiments, the activity in visual cortex when a subject imagines a visual stimulus resembles the activity elicited by a corresponding ancillary stimulus \cite{kosslyn2001neural,bartolomeo2002relationship}.

The idea of perceptual simulation can help in at least two major applications in a perceptual system. First is the internal simulation when there is a limited access to the sensory input, and second is the future prediction of the consecutive perceptual sequences. According to \cite{johnsson2009associative}, the subsystems of different sensory modalities of a multi-modal perceptual system are associated with one another. Therefore, suitable activity in some modalities that receive input can elicit activity in other sensory modalities as well. The ability of internal simulation can facilitate the activation in the subsystems of a modality even when there is no or limited sensory input but instead there is activity in subsystems of other modalities.

Moreover, in different perceptual subsystems there is a capacity to elicit continued activity even in the absence of sensory input. In other words, the perceptual process does not stop when the sensory input is blocked but rather continues by internally simulate the sequences of perceptions as proposed in the neuroscientific simulation hypothesis \cite{hesslow2002conscious}. This continuity is an important feature of a perceptual system especially for the circumstances in which the connection to the sensory input is somehow reduced.

Empirical studies of rodents' memory have also shown that the hippocampus stores and reactivates episodic memories offline \cite{Stoianov}. Based on these findings, the rodent hippocampus internally simulates neural activations (so called replays) in phases of wakeful rest during spatial navigation, as well as during subsequent sleep, which resemble sequences observed during animal's real experiences \cite{buzsaki2015hippocampal}.

Since, these internally simulated hippocampal sequences during sleeping or wakeful resting depict paths to future goals rather than only past trajectories, it is hard to believe that the hippocampus does only “replay” past experiences of a memory buffer. Therefore, it can somehow be considered in the role of planning and imagination \cite{pfeiffer2013hippocampal}; trajectories that have never been directly explored but yet reflect prior spatial knowledge \cite{liu2018generative,dragoi2013distinct} or for future event locations \cite{olafsdottir2015hippocampal}. 

To demonstrate this feature of hippocampus, a hierarchical generative model of hippocampal formation is proposed in \cite{Stoianov}, which organizes sequential experiences to a set of coherent but mutually exclusive spatio-temporal contexts (e.g., spatial maps). They have also proposed that the internally generated hippocampal sequences stem from generative replay, or the offline re-sampling of fictive experiences from the generative model.

More importantly, the continuous perceptual simulation mechanism can facilitate the anticipation of future sequences of perceptions, the prediction, that normally follows a certain perception within a modality, and also over different modalities. This occurs only if the modalities are associated in an appropriate manner. For example, a thunder light seen, would yield an anticipation of hearing a sound to be followed soon.

However, prediction of future sequences of perception is extremely important in different aspects of life such as rational decision making or social interaction. Recent developments in the conceptualization of the brain processing have shown the predictive nature of the brain, the predictive coding or predictive processing views (\cite{clark2013whatever,friston2010free}).

In fact, many perceptual phenomena can only be understood by assuming that meaningful perception is not just a matter of processing incoming information, but it is largely reliant on prior information since the brain often unconsciously and compellingly assumes (or infers) non-given information to construct a meaningful percept \cite{pezzulo2019symptom}. From a neuroscience perspective also, the theory of "predictive coding" (\cite{friston2005theory}) describes how sensory (e.g., visual) hierarchies in the brain may combine prior knowledge and sensory evidence, by continuously exchanging top-down (predictions) and bottom-up (prediction error) signals.

An important application of predictive processing lies in the recognition of others' distal intentions. This plays a substantial role in recognizing actions performed by others in advance to be prepared for planning to make an appropriate reaction for example in the case of social interactions.

A computationally-guided explanation of distal intention recognition derived from theories of computational motor control is proposed by \cite{donnarumma2017sensorimotor}. Based on the control theory \cite{pezzulo2016navigating}, proximal actions have to simultaneously fulfill the concurrent demands of first-order as well as higher-order planning. As an example, in performing action 'grasping an object', first-order planning determines object handling grasp trajectory according to immediate task demands (e.g., tuning to the orientation or the grip size for an object to be grasped) while the higher-order planning alters one’s object manipulation behavior not only on the basis of immediate task demands but also on the basis of the next tasks to be performed. Based on this view, it is necessary to simultaneously optimize proximal components of an action like reaching and grasping a bottle as well as distal components (intentions) of an action sequence such as pouring water or rather moving the bottle.

On the other hand, based on the affordance competition hypothesis \cite{cisek2007cortical}, the processes of action selection (what to do?) and specification (how to do it?) occur simultaneously and continue even during overt performance of movements. According to this hypothesis, complete action plans are not prepared for all possible actions at a given moment. There are only actions specified, which are currently available first and next many possible actions are eliminated from processing through selective attention mechanisms. Attention processing limits the sensory information that is transformed into representations of action.

Therefore, according to affordance competition hypothesis \cite{cisek2007cortical}, complete action planning is not proposed even for the final selected action. Even in cases of highly practiced behaviours, no complete pre-planned motor program or the entire desired trajectory appears to be prepared. This hypothesis emphasises on a continuous simultaneously processing of action selection and specification until the final goal is achieved.

According to \cite{vinanzi2019mindreading} there is a preliminary distinction between goal and intention. The goal represents a final desired state while the intention incorporates both a goal and an action plan to achieve it. Knowing this distinction, in both approaches presented above \cite{donnarumma2017sensorimotor,cisek2007cortical} human is involved in an ongoing process of simultaneous planning for selecting and performing the actions required to address an intention or to achieve a particular goal. 

Prediction of future sequences of state and the internal simulation of the self-or the other’s perception (reading the other’s intention) facilitates this ongoing process by providing a priori-knowledge about the future. One way to model the internal simulation of perception is to apply Associative Self-Organizing Map (A-SOM) neural networks \cite{johnsson2009associative}. This model is developed and utilized in a number of applications such as action simulation and discrimination \cite{buonamente2015discriminating,buonamente2013simulating}, letters prediction in text \cite{gil2015sarasom} and music simulation \cite{buonamente2018simulatingmusic}.

\vspace{-0.5em}
\subsection{Applications of reading intention}
\vspace{-0.5em}
There are a number of approaches performing some practical applications to somehow address the problem of robots reading intentions. Among them there is the Hierarchical Attentive Multiple Models for Execution and Recognition (HAMMER) architecture as a representative example of the generative embodied simulationist approach to understanding intentions \cite{demiris2007prediction}. HAMMER uses an inverse–forward model coupling in a dual role: either for executing an action, or for perceiving the same action when performed by a demonstrator, the imitator processes the actions by analogy with the self—‘‘what would I do if I were in the demonstrator’s shoes?’’.

Another approach proposed by \cite{dominey2011basis} is built to test the cooperation, a robotic system for cooperative interaction with humans in a number of experiments designed to play game in collaboration. A probabilistic hierarchical architecture of joint action is presented in \cite{dindo2013will}, which models the casual relations between observables (movements) and their hidden causes (goals, intentions and beliefs) in two levels: at the lowest level the same circuitry used to execute the self-actions is re-enacted in simulation to infer and predict actions performed by the partner, while the highest level encodes more abstract representations, which govern each agent’s observable behavior.

The study presented in \cite{sciutti2015investigating} proposes taking into consideration the human (and robot) motion features, which allow for intention reading during social interactive tasks. In \cite{vinanzi2019mindreading} a cognitive model is developed to perform intention reading on a humanoid partner for collaborative behavior towards the achievement of a shared goal by applying the experience.

Using intention inference to predict actions performed by others is proposed in a Computational Cognitive Model (CCM) inspired by the biological mirror neurons \cite{ang2012intent}, which applies simulation theory concept of perspective changing and how one's decision making mechanism can be influenced by mirroring other’s intentions and actions. The method presented in \cite{karasev2016intent} also predicts long-term pedestrian motions by using the discrete-space models applied to a problem in continuous space.

Recognizing plans together with the actions performed, by developing a hybrid model is proposed in \cite{granada2017hybrid}, which combines a deep learning architecture to analyze raw video data to identify individual actions and process them later by a goal recognition algorithm using a plan library to describe possible overarching activities.  Inferring and prediction of intentions for communicative purposes is done in other approaches like; the prediction of intentions in gaze-based interactions \cite{bednarik2012you}, inferring communicative intention from images \cite{joo2014visual} and predicting human intentions and trajectories in video streams \cite{xie2017learning}.

\begin{figure*}[h]
	\centering
	\includegraphics[width=\linewidth]{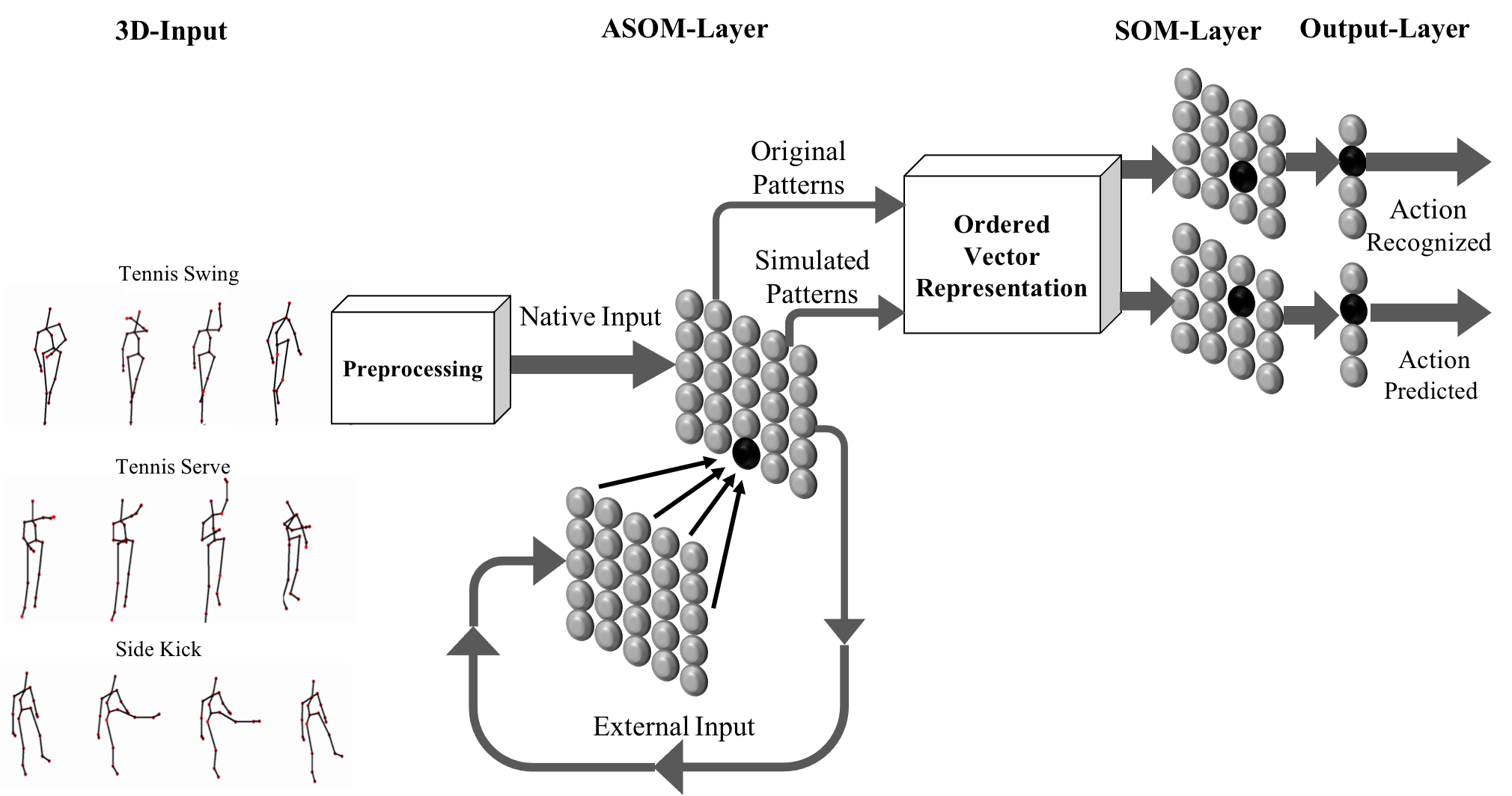}
	\caption{The architecture proposed using ASOM neural networks}
	\label{fig:asom}
\end{figure*}

\vspace{-0.5em}
\section{Proposed architecture}\label{prepro}
\vspace{-0.5em}
Action recognition architecture, figure(\ref{fig:asom}), is consists of five layers:  preprocessing layer, A-SOM layer, ordered vector representation layer, SOM layer and the output layer. Shown by figure(\ref{fig:asom}), native input is composed of consecutive 3D posture frames and external input is the activity map of A-SOM layer. It is possible to use different types of input data as the external input. For example, the main modality of a system like vision can produce the native input and other modalities like auditory or olfactory can be used to generate external inputs. Learning of A-SOM layer is done using native and external inputs.

Training A-SOM layer, original pattern vectors of action sequences are created using total activity of the network. An original pattern vector is created by connecting consecutive activation of neural map when consecutive posture frames are received as the native input by the network. Simulated pattern vectors are generated using external inputs only. To this end, partial native input is received by trained A-SOM.

The original as well as simulated patterns are next received by ordered vector representation layer to produce time invariant action pattern vectors as, which is given to SOM- and output layers. Finally, action predicted from original patterns is compared with the one predicted from simulated patterns.

\vspace{-1.0em}
\subsection{Input data and Preprocessing}
\vspace{-1.0em}
The input space is composed of 3D skeleton-based human actions sequences captured by Kinect sensor. Each dataset contains a number of action sequences while each sequence is composed of consecutive posture frames. Every posture frame contains 3D information of skeleton joints, shown in figure(\ref{fig:asom}).

Preprocessing layer uses three main functions to process input data: ego-centered coordinate transformation, scaling and attention mechanism. Ego-centered coordinate system is to make a human posture invariant of having different orientations towards the camera while acting. Scaling function used to similarly scale all posture frames to be invariant of having different distances to the camera. Attention mechanism selects skeleton joints based on their velocity, the joints moving the most play the main role in acting. A thorough description about the design and implementation of the preprocessing functions is presented in \cite{Gharaee6}. The presentation of this layer is not among the main scopes of this article.

\subsection{A-SOM layer}
\vspace{-0.5em}
A-SOM neural network of this paper is a self-organizing map (SOM), which learns to associate its native activity from the native input with a number of external activities from a number of external inputs. The input to the system at each time step $t$ is: 

\begin{equation}\label{eq:1}
X(t) = \left\{x^n(t),  x^e_1(t), x^e_2(t), x^e_3(t), ..., x^e_r(t)\right\},
\end{equation}

where $x^n(t) \in R^n$ is the native input received by the main SOM and $x^e_i(t) \in R^m$ and $1<i<r$ are the $r$ external inputs received by the $r$ number of external SOMs.

The network consists of an $I \times J $ grid of neurons with a fixed number of neurons and a fixed topology. Each neuron $n_{ij}$ is associated with $r+1$ weight vectors, among them one weight vector, $w_{ij}^n \in R^n $, is used to parametrize native input and the remaining $r$ weight vectors, $w_{ij}^{e_1} \in R^{m_1}, w_{ij}^{e_2} \in R^{m_2}, ..., w_{ij}^{e_r} \in R^{m_r}$, are used to parametrize external inputs. Weight vectors are initialized by real numbers randomly selected from a uniform distribution between 0 and 1.

At each time step, the network receives input shown by (\ref{eq:1}) and the native input of each neuron representing the distance to the input vector is calculated based on the Euclidean metric as:

\begin{equation}\label{eq:2}
z_{ij}^n(t)=||x^n(t) - w_{ij}^n(t)||^2,
\end{equation}

where $||.||$ shows $l_2$ norm. The native activity is calculated by applying an exponential function to the native input and passing it over a soft-max function computed as:

\begin{equation}\label{eq:3}
y_{ij}^n(t)=\frac{z^{s_{exp}}_{ije}}{\max\limits_{k \in I\times J} z^{s_{exp}}_{ije}},
\end{equation}

where $z_{ije}=\exp(\frac{-z_{ij}^n(t)}{\sigma})$ and $\sigma$ is the exponential factor to normalize and increase the contrast between highly activated and less activated areas, $k$ ranges over the neurons of the network grid and $s_{exp}$ shows the soft-max exponent. The external activities corresponding to the external inputs are also calculated based on the Euclidean metric:

\begin{equation}\label{eq:4}
y_{ij}^p(t)=e^{\frac{-||x^e_p(t) - w_{ij}^{e_p}||^2}{\sigma}},
\end{equation} 

where $p$ ranges over the external inputs $1<p<r$ and $\sigma$ is the exponential factor to normalize and increase the contrast between highly activated and less activated areas. Having the native activity shown by (\ref{eq:3}) and external activities shown by (\ref{eq:4}), the total activity of the network is calculated as:

\begin{equation}\label{eq:5}
Y_{ij}(t)= \frac{1}{r+1} \left(y_{ij}^n(t) + \sum_{p=1}^{p=r}{y_{ij}^p(t)}\right).
\end{equation} 

Next is to find the winning neuron $n_w$ based of the network total activity $Y_{ij}$ as:

\begin{equation}\label{eq:6}
n_{w}=\mathrm {arg} \mathrm{max}_{ij}Y_{ij}(t),
\end{equation}

where $i$ and $j$ ranges over the rows and columns of the network grid. By calculating the winner the native weights are tuned by:

\begin{equation}\label{eq:7}
w_{ij}^n(t+1)=w_{ij}^n(t)+\alpha(t)G_{ijw}(t)\left[x^n(t)-w_{ij}^n(t)\right]. 
\end{equation}

The term $0 \leq \alpha(t) \leq 1$ shows the learning adaptation strength, which starts with a value close to $1$ and decays by time as $\alpha(t) \rightarrow \alpha_{min}$ when $t \rightarrow \infty$. The neighborhood function $G_{ijw}(t)=e^{-\frac{||d_{n_w} - d_{n_{ij}}||^2}{2\rho^2(t)}}(t)$ is a Gaussian function decreasing with time, and $d_{n_w} \in R^2$ and $d_{n_{ij}} \in R^2$ are location vectors of winner $n_w$ and neuron $n_{ij}$ respectively. The term $\rho(t)$ also shows adaptation over the neighborhood radius, which starts with full length of the grid and decays by time as $\rho(t) \rightarrow \rho_{min}$ when $t \rightarrow \infty$. Thus the winner neuron receives the strongest adaptation and the adaptation strength decreases by increasing distance from the winner. As a result the further the neurons are from the winner, more weakly their weights are updated. The weights of the external inputs on the other hand are updated as the following:

\begin{equation}\label{eq:8}
w_{ij}^{e_p}(t+1)=w_{ij}^{e_p}(t)+\beta(t) x_p^e(t)\left[y_{ij}^n(t)-y_{ij}^p(t)\right],
\end{equation}

where $\beta(t)$ is a constant adaptation strength and $p$ ranges over the external inputs $1<p<r$.

\subsection{Ordered vector representation layer}
\vspace{-0.5em}

This layer is designed to create input data to SOM-layer \cite{Gharaee6} by extracting unique activation patterns of A-SOM-layer and segmenting them into the vectors having equal number of activity segments. Therefore, it conducts two operations, first the subsequent repitition of similar activations is mapped into a unique activation and then, all action pattern vectors are ordered to represent vectors of equal activity segments. 

To this end, pattern vector with the maximum number of activations is extracted and the number of its activations is calculated by $K_{max} = \max\limits_{v_n \in V}(k_{v_n})$, where $v_n$ shows one activity pattern vector, $V$ shows the set of all activity pattern vectors and $k_{v_n}$ represents the number of activations of $v_n$.

The goal is to increase the number of activations for all activity pattern vectors to $K_{max}$ by inserting new activations while preserving spatial geometry of the original pattern vectors trained by A-SOM layer. Therefore, for each activity pattern vector, the approximately optimal distance $d_{v_{n}}$ between two consecutive activations is calculated as:

\begin{equation}\label{eq:16}
d_{v_{n}}=\frac{1}{K_{max}} \left(\sum_{n=1}^{N-1} \left\Vert a_{n+1}-a_{n} \right\Vert\right),
\end{equation}

where $a_n=[x_{n}, y_{n}]$ is an activation in the 2D map and $N$ shows the total number of activations for the corresponding activity pattern vector. 

To find the location of a new insertion, the distance between a consecutive pair of activations is calculated using $\ell_{2}$ norm as $d_n=\left\Vert a_{n+1}-a_{n} \right\Vert$ and if $d_n > d_{v_{n}}$, new activation $a_p$ is inserted on the line connecting $a_{n}$ to $a_{n+1}$ with a distance of $d_{v_{n}}$ from $a_{n}$ through solving system equation of:

\begin{equation}\label{eq:18}
sys= \begin{cases}
(1): \frac{y_p-y_n}{y_{n+1}-y_n} = \frac{x_p-x_n}{x_{n+1}-x_n}\\
(2): \left\Vert a_{p}-a_{n} \right\Vert = d_{v_{n}}\\
\end{cases}
\end{equation}

where $a_{n} = [x_{n}, y_{n}]$ and $a_{n+1} =[x_{n+1}, y_{n+1}]$ and $a_p = [x_p, y_p]$.

If $d_n < d_{v_{n}}$, new activation $a_p$ is inserted on the line connecting $a_{n+1}$ to $a_{n+2}$ with a distance of $ d_{v_{n}}^{\prime}=  d_{v_{n}}-d_n$ from $a_{n+1}$ solving system equations of:

\begin{equation}\label{eq:20}
sys = \begin{cases}
(3): \frac{y_p-y_{n+1}}{y_{n+2}-y_{n+1}} = \frac{x_p-x_{n+1}}{x_{n+2}-x_{n+1}}\\
(4): \left\Vert a_{p}-a_{n+1}\right\Vert = d_{v_{n}}^{\prime}\\
\end{cases}
\end{equation}

and, $a_{n+1}$ is removed from the corresponding pattern vector. 

Insertion of new activations continues until the total number of activity segments of the corresponding pattern vector becomes equal to $K_{max}$.

\subsection{SOM layer}
\vspace{-1em}
The SOM-layer used in this article designed with a grid of $I\times J$ neurons with a fixed number of neurons and a fixed topology. Each neuron $n_{ij}$ is associated with a weight vector $w_{ij}\in{R}^n$ having the same dimension $K$ as the input vector $x(t)$. For a squared SOM the total number of neurons is the number of rows multiply by the number of columns. All elements of the weight vectors are initialized by real numbers randomly selected from a uniform distribution between 0 and 1.

At time $t$, each neuron $n_{ij}$ receives the input vector $x(t)\in{R}^n$.
The net input $z_{ij}(t)=||x(t) - w_{ij}(t)||$ at time $t$ is calculated based on the Euclidean metric. Activation of each neuron is calculated using (\ref{eq:3}) while $y_{ij}(t)=y^{n}_{ij}(t)$. Applying activity matrix $Y_{ij}(t)$, winning neuron having the strongest activation value is received using (\ref{eq:6}) and, therefore, weight vectors of all neurons $w_{ij}$ are adapted using (\ref{eq:7}).

\subsection{Output layer}
\vspace{-0.5em}
The output layer is designed as one-layer supervised neural network, which receives as its input the activity traces of the SOM-layer. The output-layer consists of a number of neurons $N$ and a fixed topology. The number $N$ is determined by the number of classes representing actions names. Each neuron $n_{i}$ is associated with a weight vector  $w_{i}\in{R}^n$. All the elements of the weight vector are initialized by real numbers randomly selected from a uniform distribution between 0 and 1, after which the weight vector is normalized, i.e. turned into unit vectors.

At time $t$ each neuron $n_{i}$ receives an input vector $x(t)\in{R}^n$. The activity $y_{i}$ of the neuron $n_{i}$ is calculated using the standard cosine metric:

\begin{equation}\label{eq:9}
y_{i}=\frac{x(t)\cdot w_{i}(t)}{||x(t)||.||w_{i}||}.
\end{equation}

During the learning phase the weights $w_{i}$ are adapted as:

\begin{equation}\label{eq:10}
w_{i}(t+1)=w_{i}(t)+\gamma x(t)\left[y_{i} - d_{i}\right],
\end{equation}

where  $\gamma$ is a constant adaptation strength and is set to 0.35. The term $y_i$ is the predicted activation by the network and $d_{i}$ is the desired activation of neuron $n_{i}$.

\section{Experiments}\label{sec:exp}
\vspace{-0.5em}
In this section the experiments showing the accuracy of the architecture are presented. To this end, three skeleton based action datasets are used to run the experiments. All settings of the system hyperparameters are shown in Table (\ref{tab:Table1}). According to Table (\ref{tab:Table1}), $\alpha$, $\beta$ and $\rho$ are calculated at each time step using:

\begin{equation}
X(t) \gets X_{d}\left(X_{f}-X(t)\right)+X(t), 
\label{eq:decay}
\end{equation}
where $X(t)$ shows the values of the changing parameters, $\alpha$, $\beta$ and $\rho$, at current time step $t$, $X_f$ shows final values of the parametrs and $X_d$ is a decaying rate.  

\begin {table}
\centering
\caption {Table shows settings of the parameters of the architecture. $\alpha$, $\beta$ and $\rho$ are initialized according to the initial values $\alpha_i$, $ \beta_i$ and $\rho_i$ and updated by decay values $\alpha_d$, $ \beta_d$ and $\rho_d$ using (\ref{eq:decay}), until they approach a final value, $\alpha_f$, $ \beta_f$ and $\rho_f$.} \label{tab:Table1}
\begin{tabular}{ |p{1.7cm}|p{2.3cm}|p{2.2cm}|}
	\hline
	\multicolumn{3}{|c|}{Hyper-Parameter-Settings} \\
	\hline
	Parameters         & A-SOM        & SOM     \\
	\hline
	Neurons          & 900     & 1600 \\
	$\sigma$         & $10^{6}$       & $10^{3}$    \\
	$s_{exp}$      & 10          & 10     \\
	$\alpha_{i}$, $\alpha_{d}$, $\alpha_{f}$    & $0.1$, $0.99$, $0.01$      & $0.1$, $0.99$, $0.01$  \\
	$\rho_{i}$, $\rho_{d}$, $\rho_{f}$    & $30$, $0.999$, $1$     & $30$, $0.999$, $1$ \\
	$\beta_{i}$, $\beta_{d}$, $\beta_{f}$    & $0.35$, $1.0$, $0.01$     & -  \\
	Metric           & Euclidean    & Euclidean \\
	Epoch           & 300    & 1500  \\
	\hline
\end{tabular}
\end{table}

\vspace{-1em}
\paragraph{Input}Input data is composed of consecutive posture frames of skeleton represented by 3D joint positions captured by a Kinect sensor shown in figure(\ref{fig:asom}). Three datasets of actions are presented in the following. For all experiments, 10-fold cross validation approach is used to split a dataset into training, validation and test sets. To this end, a test set containing 25\% of action sequences is randomly selected for each dataset. The remaining 75\% of action sequences are randomly splitted into 10 folds through which one is used for validation and the remaining are used for training the system. The best performing model on the validation set in terms of recognition accuracy is finally selected and tested on the test set.

\textit{MSRAction3D\_1 dataset} by \cite{MSR} contains 287 action sequences of 10 different actions performed by 10 different subjects in 2 to 3 different events. The actions are performed using whole body of the performer, arms as well as legs. Each posture frame contains 20 joint positions represented by 3D Cartesian coordinates. The actions are: 1.Hand Clap, 2.Two Hands Wave, 3.Side Boxing, 4.Forward Bend, 5.Forward Kick, 6.Side Kick, 7.Still Jogging, 8.Tennis Serve, 9.Golf Swing, 10.Pick up and Throw.

\textit{MSRAction3D\_2 dataset} \cite{MSR} contains 563 action sequences achieved of 20 different actions performed by 10 different subjects in 2 to 3 different events. The actions are: 1.High Arm wave, 2.Horizontal Arm Wave, 3.Using Hammer, 4.Hand Catch, 5.Forward Punch, 6.High Throw, 7.Draw X-Sign, 8. Draw Tick, 9. Draw Circle, 10.Tennis Swing, 11.Hand Clap, 12.Two Hands Wave, 13.Side Boxing, 14.Forward Bend, 15.Forward Kick, 16.Side Kick, 17.Still Jogging, 18.Tennis Serve, 19.Golf Swing, 20.Pick up and Throw.

\textit{Florence3DActions dataset} by \cite{Seidenari} contains 215 action sequences of 9 different actions performed by 10 different subjects in 2 to 3 different events. This dataset consist of 3D Cartesian coordinates of 15 skeleton joints. The actions are: 1.Wave, 2.Drink from a Bottle, 3.Answer Phone, 4.Clap Hands, 5.Tight Lace, 6.Sit down, 7.Stand up, 8.Read Watch, 9.Bow. The actions are 

Experiments are designed and implemented to present the ability to recognize human actions and to compare it with SOM architecture proposed in \cite{Gharaee5}. Next, there are illustrations showing the internally simulated action patterns compared with the original ones. Finally, the accuracy of the architecture in predicting intended actions using simulated perception is presented and compared with the results when using the original perception.

\vspace{-1em}
\paragraph{Action recognition} \label{AR}
To evaluate the ability of the system in action recognition tasks, experiments are designed using three different datasets of actions and the results are presented in Table(\ref{tab:res}). Based on the results, the system recognizes actions with about the same accuracy as SOM-architecture \cite{Gharaee1,Gharaee5,Gharaee3,Gharaee4}. However there is a slight drop in accuracy using \textit{MSRAction3D\_1} and \textit{Florence3DActions} datasets. One explanation for this reduction is when receiving external inputs together with native inputs, dimension of input data increases and, therefore, there is more complexity in learning input space.
\vspace{-1em}
\paragraph{Perception simulation}\label{PS}
Action patterns vectors are created by training A-SOM layer. Using one dimension of external input together with action inputs makes the system capable of accomplishing the internal simulation task. To evaluate this capacity, A-SOM layer was fully trained using all input resources: native and external inputs. Original patterns were created and the system received partial native input in several experiments to generate simulated patterns.

\begin{figure*}[h]
\centering
\includegraphics[width=15cm,height=7.0cm]{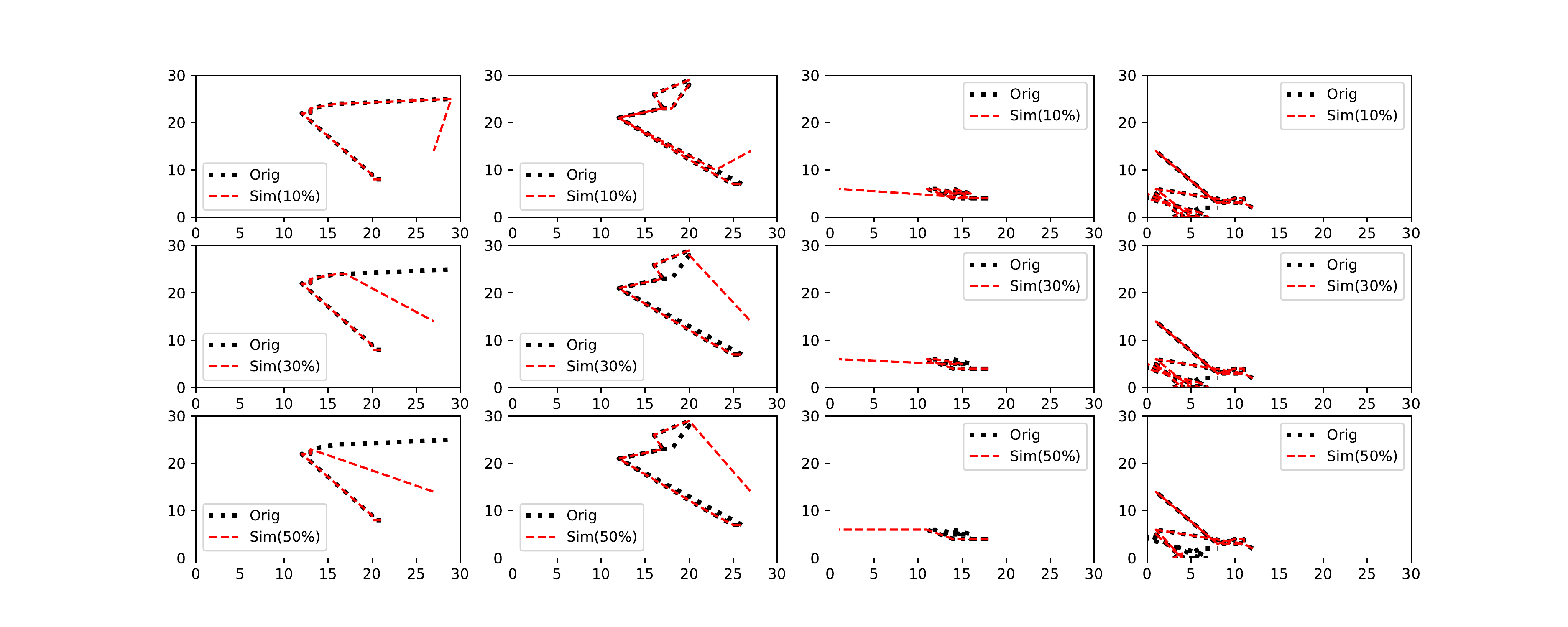}
\caption{Internal Simulation of action patterns for three different action sequences. Each row shows the comparison of internally simulated patterns of a certain percentage of data input with the original ones.}
\label{fig:patt}
\end{figure*}

Figure(\ref{fig:patt}) presents simulated and original patterns together. In each row, it shows when certain percentage of native input is deducted and replaced by zero padding, the system has to internally simulate patterns using external input only. As plots show, the system can successfully simulate action patterns similar to the original ones.

\begin{table}
	\centering
	\caption{Performance in action recognition using three different datasets of human 3D actions. The results of test experiments are presented when SOM and A-SOM neural network are applied. SOM results shows the accuracy of the framework proposed by \cite{Gharaee5}. MSR(1,2) are MSRAction3D dataset by \cite{MSR} and Florence is the Florence3DAction dataset by \cite{Seidenari}}
		{\begin{tabular}{lccc} \toprule & \multicolumn{2}{l}{Datasets} \\ \cmidrule{2-4}
				Networks & MSR(1) & MSR(2) & Florence\\ \midrule
				SOM & 86.00\% & 59.61\% & 72.22\% \\
				A-SOM & 86.50\% & 72.22\% & 74.10\% \\ \bottomrule
		\end{tabular}}
\label{tab:res}
\end{table}
\vspace{-1em}
\paragraph{Recognition using simulated perception } \label{RSP}
To evaluate the ability of the system in predicting intended actions, simulated patterns were given to SOM- and output layer in different experiments. The accuracy of predicting actions using simulated patterns is presented in Table(\ref{tab:ressim}).

\begin{figure}
	\centering
	\centerline{\includegraphics[width=9cm,height=5cm]{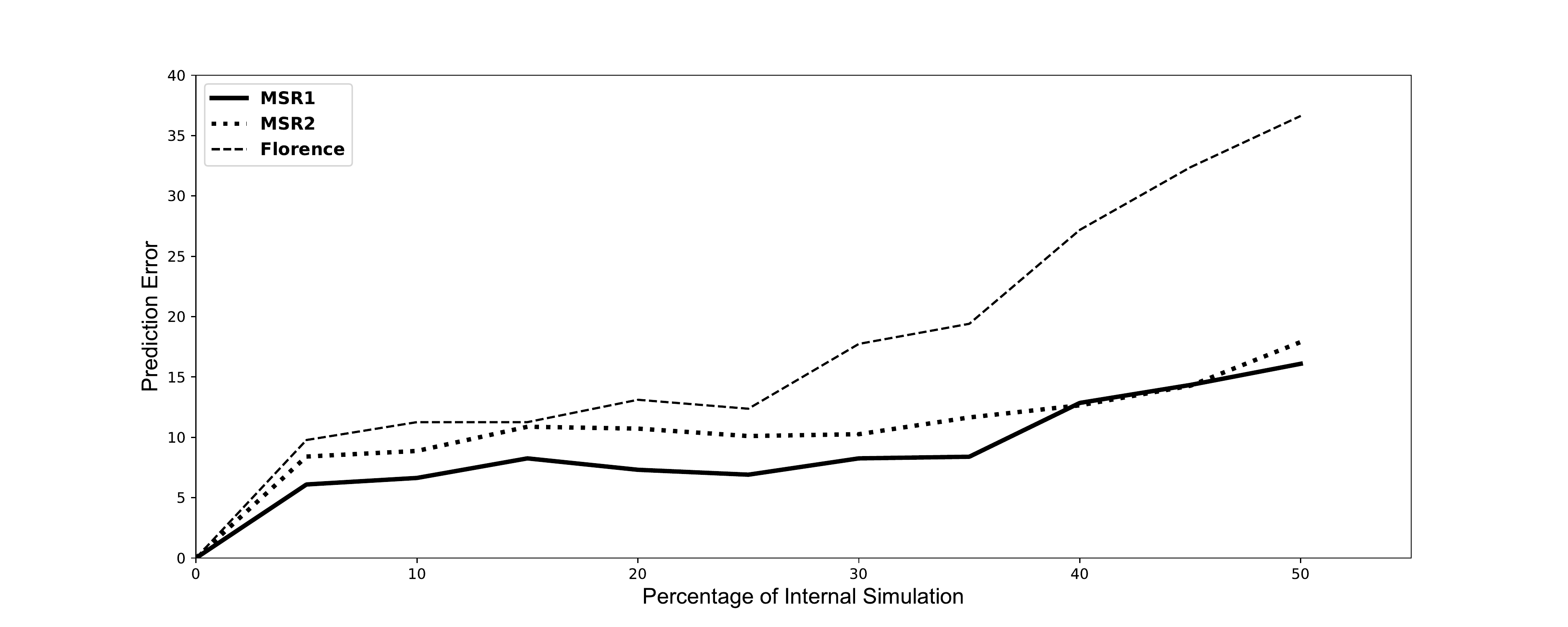}}
	\caption{Prediction errors using simulated information with three different datasets to show how much internal simulation decreases the accuracy by modifying the percentage of simulating perception. }
	\label{fig:prederr}
\end{figure}

Shown by Table(\ref{tab:ressim}), accuracy drops almost around 5\% up until simulating 35\% of action patterns. Above 35\% of simulation, performance reduction is much more significant. Prediction errors in recognizing actions when using simulation is shown in figure(\ref{fig:prederr}).

\vspace{-1em}
\section{Discussion}
\vspace{-0.5em}
The architecture proposed in this article is capable of simulating perception and applying it in predicting the intended action. This occurs by using associative self-organizing neural networks (A-SOM). Using A-SOM layer with association (external inputs) force the system to simulate the perception in the absence of native input. Three  different action 3D datasets are used in various experiments to evaluate the system performance.

Among related approaches using A-SOM neural network are those proposed by \cite{buonamente2013simulating,buonamente2015discriminating,buonamente2018simulatingmusic,gil2015sarasom}. However, in \cite{buonamente2018simulatingmusic}, the authors used A-SOM in simulating musics and in \cite{gil2015sarasom}, they focused on predicting sequences of letters using a supervised architecture based on the recurrent associative SOM (SARASOM).

Using the system presented in \cite{buonamente2013simulating,buonamente2015discriminating}, the authors have shown the A-SOM abilities in discriminating and simulating actions. The dataset used in these studies is composed of black and white contours recorded with only two human performers (Andreas and Hedlena) who perform a set of 13 different actions, one performs actions for the training set and the other for the test set.

The method proposed by \cite{buonamente2015discriminating} uses A-SOM to discriminate actions by detecting the center of activities, however no result is reported and the authors mentioned their goal as presenting investigations of self-organising principles leading to the emergence of sophisticated social abilities like action and intention recognition. Furthermore, the approaches proposed by \cite{buonamente2013simulating,buonamente2015discriminating}, have not shown when and how simulated information can be used.

In this article, the experiments are designed using three different datasets of actions \cite{MSR,Seidenari} having a larger number of sequences in comparison to \cite{buonamente2013simulating,buonamente2015discriminating} and the experiments are designed to show the abilities of A-SOM neural networks in simulating perception and applying it to accomplish a higher level task, which is predicting the intended actions. 

\begin{table*}
	\centering
	\caption{Overall test results of predicting the intended action applying varying percentages of the internally simulated action patterns using three different datasets of human 3D actions. Prediction results are presented in percentage.
	MSR(1,2) is MSRAction3D dataset by \cite{MSR} and Florence is the Florence3DAction dataset by \cite{Seidenari}}
		{\begin{tabular}{lccccccccccc} \toprule
				& \multicolumn{3}{l}{Internal simulation (\%)} \\ \cmidrule{2-12}
				Datasets &0    &5 & 10& 15&   20 &  25 & 30 & 35 & 40 & 45 & 50 \\ \midrule
				MSR(1)& 86.50 & 80.41 &   79.87 &  78.24 & 79.19 &  79.60 &  78.24 & 78.11 & 73.65  & 72.16&  70.41 \\
				MSR(2) & 72.22& 63.81& 63.35& 61.34& 61.50& 62.11& 61.96&  60.57& 59.57& 57.95& 54.32 \\
				Florence & 74.10& 72.22 &    70.74 &  70.74 &68.89 &  69.63 &  64.26 & 62.59 & 54.81  & 49.63&  45.37\\ 
				\bottomrule
		\end{tabular}}
	\label{tab:ressim}
\end{table*}

Other related approaches address the problem of robots reading intentions by designing experiments based on a collaborative task \cite{dominey2011basis,vinanzi2019mindreading,demiris2007prediction}, such as playing game between a human and a robotic arm \cite{dominey2011basis}, or a humanoid \cite{vinanzi2019mindreading}. In these studies reading the intentions occurs through continuous collaboration, like by asking questions. While in the experiments of this study the system applies its learned knowledge to make simulations and uses the simulated information for the prediction.

In other approaches using images or video streams to predict human intentions and trajectories in video streams \cite{xie2017learning} and to recognize activities and plans \cite{granada2017hybrid}, the focus is to read intentions in activities rather than actions, while the approach proposed in this article concentrates on predicting the intended actions to be performed using only the motion trajectories. Moreover, the study presented in this article does not make use of other entities or objects to recognize or predict the intention.
\vspace{-1.7em}
\section{Conclusion}
\vspace{-1.3em}
In conclusion, associative self-organizing maps are employed in a cognitive architecture to predict the intended action using internal simulation of perception in the absence of native input. Running several experiments using three different datasets of actions show the accuracy of the proposed framework in recognizing and predicting the intended actions. In future, developing the architecture applying information of prediction and objects involved in action to improve segmention of a sequence recognized online \cite{Gharaee2,Gharaee7} is crucial.      

\vspace{-1em}
\section*{Acknowledgement(s)}
\vspace{-0.5em}
This work was partially supported by the Wallenberg AI, Autonomous Systems and Software Program (WASP) funded by the Knut and Alice Wallenberg Foundation.

\bibliographystyle{apalike}
\bibliography{References}

\end{document}